\documentclass[letterpaper, 10 pt, conference]{ieeeconf}  
\IEEEoverridecommandlockouts                              

\usepackage{amsmath} 
\usepackage{gensymb}
\usepackage{textcomp}
\usepackage{mathtools}
\usepackage{xargs}  
\usepackage{units}
\usepackage{pgfplots}
\usepackage[]{caption}
\usepackage{floatrow}   
\usepackage{subcaption}
\usepackage{graphicx,xcolor} 
\usepackage{bm}
\usepackage{siunitx}
\usepackage{booktabs}
\usepackage{siunitx}
\usepackage{pgfplots}
\usepackage{colortbl}
\usepackage{color}
\usepackage{multicol} 
\usepackage[export]{adjustbox}
\pgfplotsset{compat=newest}
\usepackage{layouts}
\usepgfplotslibrary{groupplots}
\usepackage[font=small,labelfont=bf,tableposition=top]{caption}

\usepackage{xspace}
\newcommand*{\eg}{e.g.\@\xspace}
\newcommand*{\ie}{i.e.\@\xspace}

\makeatletter
\newcommand*{\etc}{%
    \@ifnextchar{.}%
        {etc}%
        {etc.\@\xspace}%
}
\makeatother

\usepackage[bordercolor=white,backgroundcolor=gray!30,linecolor=black,colorinlistoftodos]{todonotes}
\newcommandx{\timo}[1]{\todo[color=yellow, inline]{timo: {#1}}}
\newcommandx{\igor}[1]{\todo[color=green, inline]{igor: {#1}}}
\newcommandx{\cesar}[1]{\todo[color=lightgray, inline]{cesar: {#1}}}
\newcommandx{\tj}[1]{\todo[color=red, inline]{tj: {#1}}}

\newlength\figureheight
\newlength\figurewidth

\makeatletter
\let\NAT@parse\undefined
\makeatother
\usepackage{hyperref}

\title{\LARGE \bf
Flexible Stereo: Constrained, Non-Rigid, Wide-Baseline Stereo Vision for Fixed-Wing Aerial Platforms}

\author{Timo Hinzmann, Tim Taubner, and Roland Siegwart
\thanks{All authors are with the Autonomous Systems Lab, ETH Zurich,
Leonhardstrasse 21, LEE,
CH-8092 Zurich, Switzerland.
{\tt \{\underline{firstname.lastname}}\}@mavt.ethz.ch.}
}

\usepackage{xcolor}

\newcommand\Ione{\mathbf{I_1}}
\newcommand\Itwo{\mathbf{I_2}}

\newcommand\va{\bm{a}}
\newcommand\vq{\bm{q}}

\newcommand\vp{\bm{p}}

\newcommand\vn{\bm{n}}
\newcommand\vv{\bm{v}}
\newcommand\vx{\bm{x}}
\newcommand\vy{\bm{y}}
\newcommand\vz{\bm{z}}
\newcommand\vC{\bm{C}}
\newcommand\vD{\bm{D}}
\newcommand\vF{\bm{F}}

\newcommand\vI{\bm{I}}
\newcommand\vG{\bm{G}}

\newcommand\vP{\bm{P}}
\newcommand\vQ{\bm{Q}}

\newcommand\vT{\bm{T}}

\newcommand\vom{\bm{\omega}}
\newcommand\vmu{\bm{\mu}}
\newcommand\vsi{\bm{\sigma}}
\newcommand\vSi{\bm{\Sigma}}
\newcommand\vth{\bm{\theta}}

\DeclareMathOperator{\diag}{diag}

\begin{document}

\maketitle
\thispagestyle{empty}
\pagestyle{empty}

\begin{abstract}
This paper proposes a computationally efficient method to estimate the time-varying relative pose between two visual-inertial sensor rigs mounted on the flexible wings of a fixed-wing unmanned aerial vehicle (UAV).
The estimated relative poses are used to generate highly accurate depth maps in real-time and can be employed for obstacle avoidance in low-altitude flights or landing maneuvers.
The approach is structured as follows:
Initially, a wing model is identified by fitting a probability density function to measured deviations from the nominal relative baseline transformation.
At run-time, the prior knowledge about the wing model is fused in an Extended Kalman filter~(EKF) together with relative pose measurements obtained from solving a relative perspective N-point problem (PNP), and the linear accelerations and angular velocities measured by the two inertial measurement units (IMU) which are rigidly attached to the cameras.
Results obtained from extensive synthetic experiments demonstrate that our proposed framework is able to estimate highly accurate baseline transformations and depth maps.
\end{abstract}
\section{INTRODUCTION}
Reliable and long-range obstacle detection is essential to enable low-altitude flights or landing maneuvers for fixed-wing unmanned aerial vehicles (UAVs).
While full size drones often employ precise Lidar or Radar systems, small-scale UAVs usually cannot afford to carry this type of heavy payload with high power consumption, but need to rely on cheaper and more light-weight optical camera systems.
To recover the absolute scale information from cameras, either a classical stereo setup with a fixed and calibrated baseline transformation is used, or virtual stereo pairs are computed from a monocular camera setup.
\subsection{Monocular Camera Setup: Flying into the Epipole}
Monocular visual-inertial sensor setups have proven to be well suited for applications such as simultaneous localization and mapping (SLAM) \cite{Klein2007} or planar dense reconstruction \cite{Fusiello2000}.
The algorithms perform best when using a down-looking camera where the epipoles are outside of the camera field of view and fronto-parallel motion with respect to the ground is performed.
However, within the application of obstacle detection and avoidance, the region of interest is in front of the aircraft and a forward facing or oblique camera needs to be employed.
In this setup, the optical axis is closely aligned with the aircraft's direction of flight, resulting in a blind spot just in the region of interest.
While this problem can be partially overcome by using, for instance, polar rectification \cite{Pollefeys1999}, the procedure is inherently error prone since the pixels close to the image center have only small changes in image coordinates in the subsequent frame.
\begin{figure}[t]
\includegraphics[width=\linewidth, trim=44 8 0 20,clip=true]{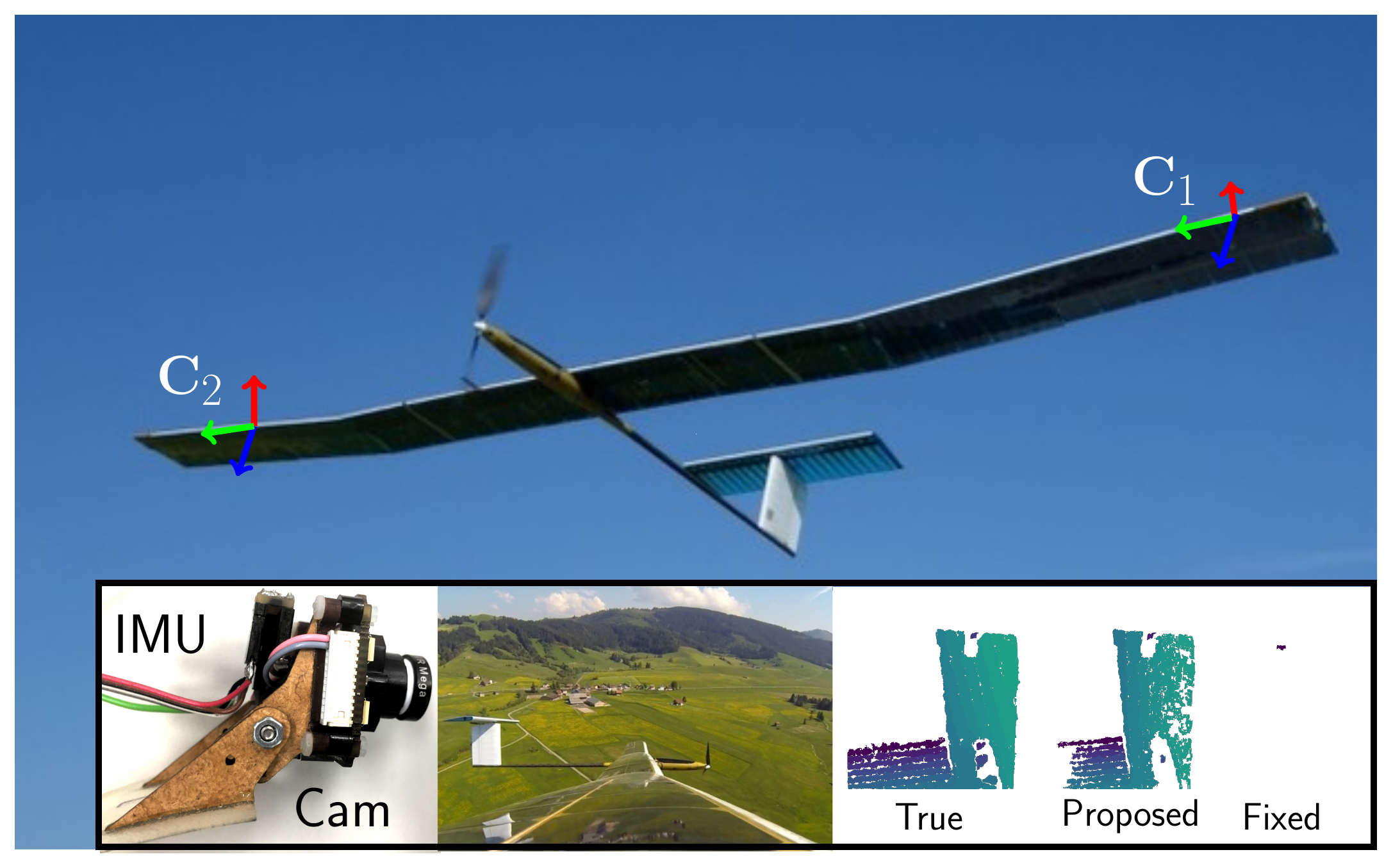}
\caption[Caption for LOF]{Envisioned use-case for the proposed non-rigid wide-baseline stereo vision algorithm for fixed-wing UAVs.\footnotemark \ Bottom left: One of the two light-weight, rigid camera-IMU rigs. Bottom right: Characteristic frame from synthetic dataset comparing depth map obtained from ground-truth poses, estimated poses, and poses obtained from a fixed baseline assumption.}
\label{fig:teaser}
\end{figure}
\footnotetext{A video illustrating \emph{AtlantikSolar}'s flapping wing behavior is accessible under \url{https://youtu.be/8m76Mx9m2nM}.}

\subsection{Fixed-Baseline Stereo}
Due to the shortcomings of the monocular sensor setup,  rotary-wing UAVs, such as quadro- or hexacopters, are usually equipped with a fixed baseline stereo pair to perform onboard and real-time obstacle avoidance while providing a high level of reliability.
However, the baseline of the stereo rigs, and hence the range where low depth uncertainties are obtained, are optimized for indoor scenarios with maximum speeds of few meters per second.
The small baselines, ranging from $\unit[9]{mm}$ up to $\unit[0.5]{m}$ \cite{Warren2016}, leave not enough time for short-term path planning, control, and actuators to react in highly dynamic fixed-wing UAV scenarios.
Upscaling the baseline to more suited depth ranges while still ensuring rigidity of the stereo rig, on the other hand, would impair aerodynamics, the required flexibility of the wing, and inevitably lead to a higher payload mass.
\subsection{Aim of this Work: Non-Rigid Wide-Baseline Stereo}
Consequently, for fixed-wing aircrafts it seems appealing to take advantage of the wing span and to mount the cameras on the outer wing region.
However, especially solar-powered aircrafts \cite{Oettershagen2017} which are optimized for wing area and updraft, show flapping wing behavior leading to translational and rotational deformations  from the nominal baseline in the range of several degrees and centimeters. 
These disturbances affect the depth estimation in two ways:
Firstly, since the assumption of an accurately calibrated fixed relative transformation between the stereo pair is violated, the depth estimates become heavily distorted.
For example, disturbances in yaw and pitch angle estimates lead to errors in the depth estimate which are quadratic to the depth \cite{Dang2009}.
Secondly, stereo vision algorithms usually rely on matching correspondences along epipolar lines \cite{Hartley2004}.
Therefore, even small errors in the estimated rotation can already be fatal since the correspondences are searched along the wrong line.
This leads to an almost empty depth map since no correspondences can be found.
It could be overcome by a full 2D search over the whole image which, however, is not computationally feasible in real-time.
The aim of this work is therefore to estimate the time-varying relative pose between the two camera-IMU rigs with low delay and close to the image capture frequency as illustrated in Fig. \ref{fig:teaser}.
We achieve this by fusing the high-frequency, low-variance inertial sensors with the low-frequency, bias-free vision-based relative pose estimation in combination with the probabilistic wing model in an efficient Extended Kalman Filter (EKF) formulation \cite{Achtelik2011,Achtelik2014}.

In summary, we present a light-weight formulation for wide-baseline non-rigid stereo that builds up on \cite{Achtelik2011}.
Compared to \cite{Achtelik2011}, we see the following contributions:
Most importantly, we propose a wing model in form of a relative pose prior.
While \cite{Achtelik2011} makes no assumption about the relative pose of the IMUs, we take advantage of our wing deflection model to better constrain the EKF and to identify and reject visual outliers.
While \cite{Achtelik2011} treated the vision module as a black box, we extract and match features and incorporate a relative perspective n-point (PNP) solver as vision-based relative motion estimation module.
The pipeline is validated extensively based on synthetic datasets, in particular, with respect to the quality of a) the estimated baseline transformation, and b) the resulting depth map.
\section{RELATED WORK}
The \textit{auto-calibration} problem is the retrieval of the rigid, non-changing stereo baseline transformation in known \cite{furgale2013unified} or unknown \cite{warren2013online,hansen2012online} environments.
In the latter, images are taken over time and keypoints are extracted to perform stereo bundle adjustment.
This can be extended towards \textit{re-calibration} of stereo rigs during operation, also known as \textit{online-calibration}.
By having a good initial estimate, \eg obtained from auto-calibration, the baseline can be re-estimated assuming it only changed slightly.
The techniques are then essentially the same as for auto-calibration in unknown environments.
Perturbations from the nominal stereo baseline can be detected by an increase in reprojection errors, hence triggering a re-calibration process.
Extending further on this idea, Warren et al. \cite{warren2013online} continuously estimate slow changes (\eg thermally induced) in the baseline.
The key difference of the \emph{continuous self-calibration} methods are that they do not assume that the baseline is rigid over the whole duration of the calibration process.
However, in contrast to our setting, the baseline changes in \cite{warren2013online} are still assumed to be only small and slow.
Work on continuous calibration in presence of high-frequent noise exists.
In \cite{warren2016long}, a down-looking stereo pair with a wide-baseline (\SI{0.7}{m}) is employed and the relative transform between the cameras, together with the poses of the stereo rig itself is estimated offline in a bundle adjustment problem.
Similarly to our work, Warren et al. incorporate prior knowledge of the deformation in the stereo rig.
This is achieved in form of a cost-based bound to tightly constrain the estimated transform.
However, in \cite{warren2016long}, the anticipated changes in baseline are induced by vibration and hence relatively small (a few millimeters in translation and only tenth of a degree in rotation) making the prior a good assumption.
In contrast, we are dealing with an increase of two order of magnitudes (decimeters and more than ten degrees).\newline
We are aware of two different approaches to real-time baseline estimation in which the baseline deviations closer resemble our scenario:
The \emph{first approach}, described in \cite{lanier2011modal, lanier2010, Short2009a}, is based on a modal-analysis of a wing which enables to estimate vibrational disturbances by only measuring with accelerometers (and no gyroscopes).
In \cite{lanier2011modal}, this method is introduced and show-cased on an Euler-Bernoulli beam model employing two accelerometers. 
This is extended in \cite{lanier2010} to wing-mounted stereo rigs and experimental results are obtained during periodic and random excitation of a real wing model.
A total of 24 accelerometers are mounted over the wing to perform modal analysis.
The camera mount-points are on the wing-tips resulting in a theoretical baseline of \SI{240}{cm}.
However, only an analysis of the wing and the corresponding baseline estimation errors was performed.
The results suggest significant improvements over a fixed-baseline assumption.
Note that these results are obtained in experiments performed in small scale and then extrapolated.
In contrast, we mount the IMUs (including a gyroscope) spatially close and rigidly to the cameras and do not rely on a physically motivated wing-model.
Furthermore, their approach suggests taking images when the motion is minimal to minimize effects of motion blur.
Instead we rely on global shutter cameras with short and synchronized shutter times.

The \emph{second approach} is described in \cite{Achtelik2011}:
It employs an EKF to estimate the relative position between two micro air vehicles (MAV), each equipped with an IMU and a down-looking camera, assuming an overlapping field of view.
The authors leave the visual part as a black box and suggest the use of a visual SLAM framework with scale propagation such as PTAM \cite{Klein2007}.
In contrast to our work, since MAVs can move independently of each other, no prior knowledge on the baseline transform is assumed.
Our work is based to a large extent on the EKF described in \cite{Achtelik2011}, but extends the measurement model by fusing our prior knowledge of the baseline. Furthermore, the vision module is implemented in form of a relative PNP solver.
Since the latter is based solely on sparse correspondence matching of the current stereo pair, no scale propagation can be performed which simplifies the EKF.
Furthermore, as the relative baseline transformation may change quickly during flight, no feature tracking or descriptor matching over time is performed.
\section{METHODOLOGY}
\begin{figure}[bth]
\includegraphics[width=1\linewidth]{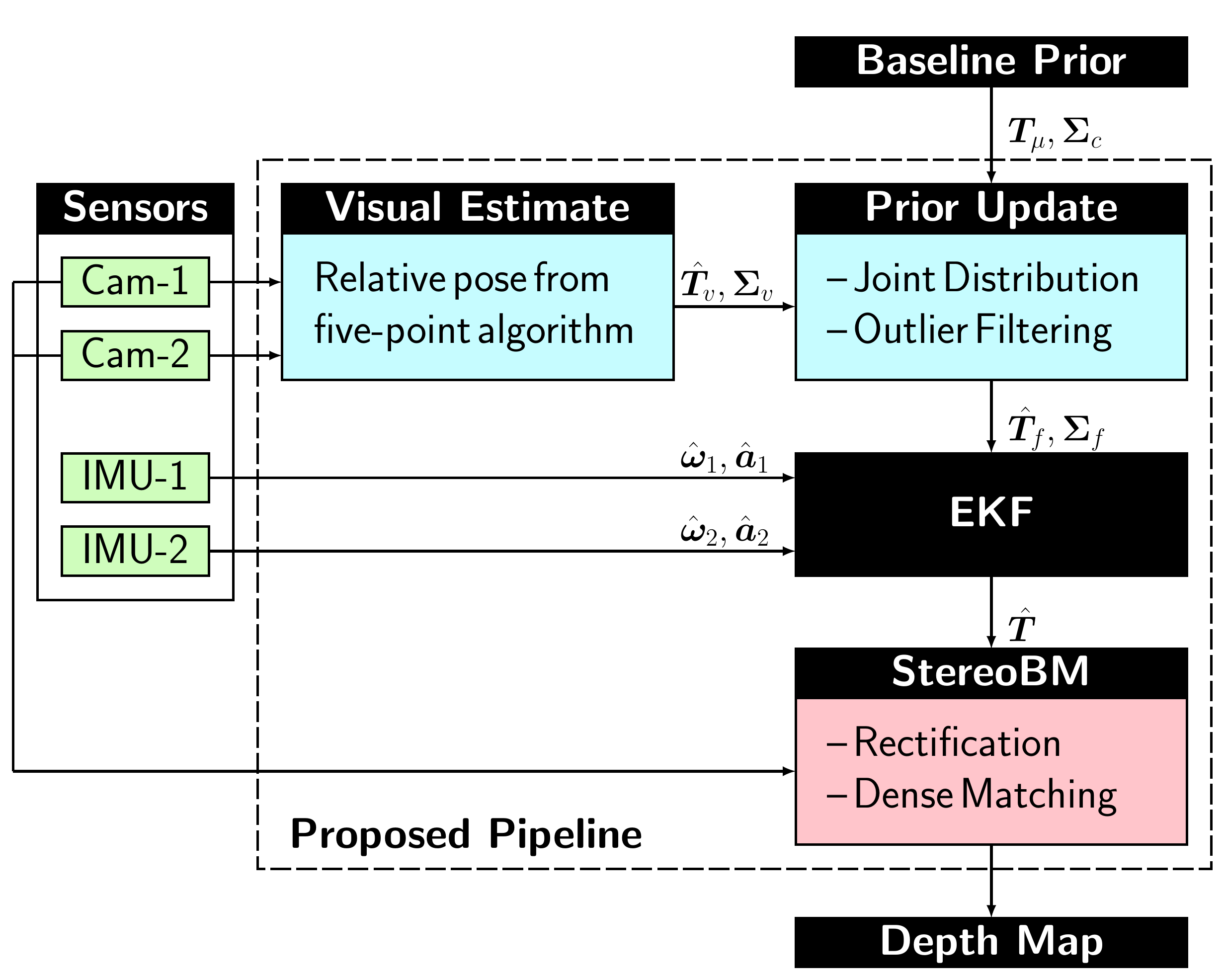}
\caption{Proposed framework to estimate the time-varying relative transformation between two cameras. The efficient EKF formulation fuses a wing deformation prior with the information from two IMUs and two cameras.}
\label{fig:overview}
\end{figure}

An overview of our proposed framework is illustrated in Fig. \ref{fig:overview}.
The incoming image stream is used both, for visual estimation of the relative pose using a relative PNP solver, and for generating depth maps.
The visual estimates are fused with the baseline prior and fed into an EKF together with high-frequent ($\geq 100$ Hz) IMU measurements.
The obtained filtered estimate of the  baseline transformation $\hat \vT$ is then used to rectify the images so that dense matching can be done reliably and fast via correspondence search along epipolar lines \cite{hartley2003multiple}.
The resulting depth maps can be fed into external applications, \eg a flight controller to avoid obstacles, or to reconstruct the environment.
We adopt the EKF as described in \cite{Achtelik2011}.
There is one fundamental modification:
As our visual pose estimator is based solely on the most recent two frames and does not have scale propagation we do not include $\lambda$ in our state vector.
Correspondingly, our measurement model for obtaining relative pose estimates is adopted.
For completeness and to clarify the modifications with respect to \cite{Achtelik2011}, we summarize all elements of the Kalman filter that are required for understanding in the following.
\subsection{The State Vector}
The state consists of the relative angular rates $\vom_1^1$, $\vom_2^2$, the linear accelerations expressed in the world frame $_w \va_1^1$, $_w \va_2^2$, the relative orientation expressed as a quaternion $\bar q_1^2$ and the (metric) relative position $\vp_1^2$ and finally the relative velocity $_w \vv_1^2$.
Stacked together these values form the 22-element state vector
\begin{equation}
 \vx = \left[ \mbox{$\bar q_1^2$}^\top \; {\vom_1^1}^\top \; {\vom_2^2}^\top \; {\vp_1^2}^\top \; {_w \vv_1^2}^\top \; {_w \va_1^1}^\top \; {_w \va_2^2}^\top \right]^\top.
\end{equation}
To simplify notation, super- and subscripts are dropped in the following and the states are simply denoted by 
\begin{equation}
 \vx = \left[ {\bar q}^\top \; \vom_1^\top \; \vom_2^\top \; \vp^\top \; \vv^\top \; \va_1^\top \; \va_2^\top \right]^\top.
\end{equation}
\subsection{State Equations}
Angular velocities $\vom_1, \vom_2$ and linear accelerations $\va_1$, $\va_2$ are modeled as independent random zero-mean Gaussian walks with (diagonal) covariances $\vsi_{\vom_1}, \vsi_{\vom_2}, \vsi_{\va_1}$ and $\vsi_{\va_2}$.
The corresponding random vectors are denoted by $\vn_{\vom_1}$, $\vn_{\vom_2}$, $\vn_{\va_1}$ respectively $\vn_{\va_2}$.
For quaternion multiplication the angular velocities are represented as quaternions $\bar \omega_i = [0 \; \vom_i^\top]^\top$ (for $i = 1,2$), $\vC$ denotes the rotation matrix corresponding to the quaternion $\bar q$ and $\lfloor \vom_1 \times \rfloor$ denotes the skew-symmetric matrix.
The final state equations are as follows:
\begin{align}
\dot{\bar q} & = 0.5 \cdot \left({\bar q} \otimes \bar \omega_2 - \bar \omega_1 \otimes \bar q \right) \label{update:q} \\
\dot\vom_1 & = \vn_{\vom_1} \\
\dot\vom_2 & = \vn_{\vom_2} \\
\dot\vp & = \vv - \lfloor\vom_1 \times \rfloor \cdot \vp \label{update:p} \\
\dot\vv & = \vC \cdot \va_2 - \va_1 - \lfloor \vom_1 \times \rfloor \cdot \vv \label{update:v} \\
\dot\va_1 & = \vn_{\va_1} \\
\dot\va_2 & = \vn_{\va_2}
\end{align}
Detailed derivations can be found in \cite{Achtelik2011}.
\subsection{Error State Representation}
\label{ekf:errstate}
From here on, an estimated state is denoted by $\hat .$ and a preceding $\Delta$ denotes the error state for an additive error.
For the quaternion $\bar q$ a multiplicative error model is used.
The error is denoted by $\delta\vth$.
The error state vector is then
\begin{equation}
\tilde \vx = \left[ \delta\vth^\top \; \Delta\vom_1^\top \; \Delta\vom_2^\top \; \Delta\vp^\top \; \Delta\vv^\top \; \Delta\va_1^\top \; \Delta\va_2^\top \right]^\top.
\end{equation}
Again, we refer to \cite{Achtelik2011} for derivations of the error state update equations.
Noting that $\hat \vC$ is the corresponding rotation matrix of the quaternion $\hat \vq$, the final error state equations are
\begin{align}
\dot{\delta\vth} = &\ -\lfloor \hat\vom_2 \times \rfloor \cdot \delta\vth - {\hat\vC}^\top \cdot \Delta\vom_1 + \Delta\vom_2\\
\dot{\Delta\vom_1} = &\ \vn_{\vom_1} \\
\dot{\Delta\vom_2} = &\ \vn_{\vom_2} \\
\dot{\Delta\vp} = &\ \lfloor \hat\vp \times \rfloor \cdot \Delta\vom_1 - \lfloor\hat\vom_1\times\rfloor\cdot\Delta\vp + \Delta\vv \\
\dot{\Delta\vv} = &\ -\hat\vC\cdot\lfloor\hat\va_2\times\rfloor\cdot\delta\vth + \lfloor\hat\vv\times\rfloor\cdot\Delta\vom_1 \nonumber \\
   &\ - \lfloor\hat\vom_1\times\rfloor\cdot\Delta\vv - \Delta\va_1 + \hat\vC\cdot\Delta\va_2 \\
\dot{\Delta\va_1} = &\ \vn_{\va_1} \\
\dot{\Delta\va_2} = &\ \vn_{\va_2}
\end{align}
\subsection{State Covariance Prediction}
These equations enable the computation of the continuous system matrix $\vF_c = \frac{\partial\dot{\tilde\vx}}{\partial\tilde\vx}$ and noise matrix $\vG_c = \frac{\partial\dot{\tilde\vx}}{\partial\vn}$ with $\vn = \left[ \vn_{\vom_1}^\top \; \vn_{\vom_2}^\top \; \vn_{\va_1}^\top \; \vn_{\va_2}^\top \; \right]^\top$.
The computed Jacobi matrices are given in \cite[(4.21)]{Achtelik2014}.
Assuming $\vF_c$ to be constant over the integration period yields $\vF_d = \exp(\Delta t \cdot \vF_c)$ for a given time-step $\Delta t$.
This is approximated to the zeroth order term by expanding the exponential series, resulting in $\vF_d \approx \vI + \vF_c\Delta t$.
With the continuous time noise covariance matrix $\vQ_c = \diag\left(\left[\vsi^2_{\vom_1} \; \vsi^2_{\vom_2} \; \vsi^2_{\va_1} \; \vsi^2_{\va_2}\right]\right)$, the discrete time noise covariance matrix $\vQ_d$ is computed according to \cite{Maybeck1979}.
Assuming $\vF_d$ to stay constant during the integration period yields
\begin{align}
\vQ_d & = \int_0^{\Delta t}\vF_d(\tau) \cdot \vG_c \cdot \vQ_c \cdot \vG_c^\top \cdot \vF_d(\tau)^\top \cdot d\tau \\
      & = \Delta t \cdot \vF_d \cdot \vG_c \cdot \vQ_c \cdot \vG_c^\top \cdot \vF_d^\top.
\end{align}
The updated state covariance matrix for the EKF is then computed as
\begin{align}
\vP_{k+1 | k} = \vF_d \cdot \vP_{k|k}\cdot\vF_d^\top + \vQ_d.
\end{align}
\subsection{State Prediction}
The states are predicted by zeroth order integration according to (\ref{update:q}), (\ref{update:p}), respectively (\ref{update:v}) given above:
\begin{align}
\hat{\bar q}_{k+1} & = \hat{\bar q}_{k} + 0.5 \cdot \Delta t \cdot (\hat{\bar q}_k \otimes \hat{\bar\omega}_{2,k} - \hat{\bar\omega}_{1,k} \otimes \hat{\bar q}_k) \\
\hat\vp_{k+1} & = \hat\vp_k + \left(\hat\vv_k - \lfloor\hat\vom_{1,k}\times\rfloor\cdot\hat\vp_k\right)\cdot\Delta t\\
\hat\vv_{k+1} & = \hat\vv_k + \left(\hat\vC\cdot\hat\va_{2,k} - \hat\va_{1,k} - \lfloor\hat\vom_{1,k}\times\rfloor\cdot\hat\vv_k\right)\cdot\Delta t
\end{align}
\subsection{Vision-Based Relative Pose Measurements}
The vision-based relative pose estimates are obtained in three steps.
Classical feature descriptor, matcher, and relative PNP types are selected for a proof of concept:
Firstly, SURF keypoints \cite{Bay2006} are detected and their descriptors are extracted in both images.
Secondly, feature correspondences are established using a FLANN-based matcher \cite{Muja2009}.
Finally, the bearing vectors are computed for both frames and every matched feature.
Based on the two sets of bearing vectors, a relative pose up to a scale is estimated by means of finding the fundamental matrix $\vF$ \cite{Hartley2004}.
The estimated fundamental matrix $\vF$ is then converted to a unit vector $\hat\vp_v$ representing the direction of translation and the rotation quaternion $\hat{\bar{q}}_v$.
To solve the PNP problem, we employ the 5-point Nister algorithm \cite{Nister2004}.
Since PNP solvers are sensitive to wrong associations, Random Sample Consensus (RANSAC) \cite{Fischler1981} is employed \cite{Kneip2014}.
To fuse the visual estimates, the output of the algorithm is interpreted probabilistically: 
The measurements are modeled as 
\begin{align}
\hat \vp_v &= \frac{\vp + \Delta\vp_v}{\left\| \vp + \Delta\vp_v \right\|_2}  \\ 
\hat{\bar q}_v &= \bar q \otimes \delta\bar q_v
\end{align}
where $\hat\vp_v$ and $\hat{\bar q}_v$ are the direction of translation respectively orientation estimated by the vision-based relative pose estimation module (i.e. in our case the PNP solver). The corresponding measurement error is denoted by $\Delta\vp_v$, $\delta\bar q_v$.
Approximating the error quaternion $\delta\bar q_v$ by small angles $\delta\vth_v$ allows expressing the error transformation as a 6-dimensional vector $\delta\vT_v = \left[ \delta\vth_v^\top \; \Delta\vp_v^\top \right]^\top$.
This error is then approximated by a zero-mean Gaussian with covariance $\vSi_v$.
\subsection{Establishing a Wing Model}\label{baseline-prior}
In this section, the notion of baseline calibration is extended beyond rigidity.
Instead of finding the fixed relative transform between the two cameras of a stereo rig as the result of an (usually over-constrained) optimization problem flexibility is embraced.
The wing model is captured by a nominal baseline transformation $\vT_\mu$ and a probabilistic error model captured in a random vector $\delta\vT$.
For notational clearance the baseline transform $\vT_\Ione^\Itwo$ is simply denoted by $\vT$ or in its parts $\bar q$ and $\vp$.
The mean transform $\vT_\mu$ is expressed in parts by the mean quaternion $\bar q_\mu$ and the mean relative position $\vp_\mu$.
This allows to express the disturbance in rotation via a multiplicative error model $\bar q = \bar q_\mu \otimes \delta\bar q$ and in position as a simple additive error $\vp = \vp_\mu + \Delta\vp$.
The (small) quaternion $\delta \bar q$ is then approximated via the small angle approximation $\delta\vth$.
With that, $\delta \vT = \left[ \delta\vth^\top \; \delta\vp^\top \right]^\top$ is defined.
Note that the random vector $\delta \vT$ has zero mean.
In the spirit of the EKF, the error of the \textit{calibration} is modeled as a Gaussian: $\delta \vT \sim \mathcal{N}(\bm 0, \vSi_c)$.
The six dimensions of $\delta \vT$ are correlated (i.e. when there is a roll-disturbance, there is also a disturbance in the z direction), however not linearly.
Therefore, the distribution is approximated as independent Gaussians and thus the covariance matrix is diagonal: $\vSi_c = \diag(\vsi_{\delta\vT}^2)$.
\subsection{Fusing the Baseline Prior with Visual Estimates}
\label{combpos}
The visual measurements of the baseline $\hat{\bar q}_v$ and $\hat \vp_v$ are then combined with the baseline prior to obtain the maximum a posteriori estimates $\hat{\bar q}_f$ and $\hat \vp_f$.
Since both, the baseline prior and the measurement error, are modeled as zero-mean Gaussians, the a posteriori distribution is again a Gaussian.

The visual estimates are expressed as the deviations $\delta\hat{\bar q}_v$ from the calibrated mean such that $\hat {\bar q}_v = \bar q_\mu \otimes \delta\hat{\bar q}_v$ and $\Delta\hat\vp_v = \hat \vp_v - \vp_\mu$.
The error quaternion $\delta\hat{\bar q}_v$ is approximated by small angles $\delta\hat\vth_v$.

The new estimates of the baseline deviation $\delta\hat{\bar q}_f$ and $\Delta\hat\vp_f$ are computed such that the estimated baseline is $\hat{\bar q}_f = \bar q_\mu \otimes \delta\hat{\bar q}_f$ and $\hat\vp_f = \vp_\mu + \Delta\hat\vp_f$.
Again, $\delta\hat{\bar q}_f$ is approximated by small angles $\delta\hat\vth_f$.
By interpreting the fusion of the measurement with the prior as a KF update step, the new estimates and the a posteriori covariance matrix is given by:
\begin{align}
\begin{bmatrix}\delta\hat\vth_f \\ \Delta\hat\vp_f\end{bmatrix} & = \vSi_c(\vSi_c + \vSi_v)^{-1}\begin{bmatrix}\delta\hat\vth_v \\ \Delta\hat\vp_v \end{bmatrix} \\
\vSi_f & = \vSi_c - \vSi_c(\vSi_c + \vSi_v)^{-1}\vSi_c
\end{align}
\subsection{Visual Outlier Rejection}
The baseline prior is used to filter outlier estimates from the relative PNP pipeline.
Depending on the scene, the visual pose estimation may give severe outliers.
This happens, for instance, if the cameras see mostly uniformly colored sky or landscape without salient keypoints for detection and matching.
In these cases, the visual pose estimate is ignored altogether and, instead, an artificial measurement update corresponding to the baseline prior is fed into the EKF.
Filtering of outliers is achieved by limiting the maximum distance of the visual estimate $\hat\vT_v$ to the mean baseline $\vT_\mu$ in any dimension by a factor $k$ (in our case 2) of the standard deviation.
That is, we set $\hat{\bar q}_f = \hat{\bar q}_\mu$ and $\hat \vp_f = \hat \vp_\mu$ if $\left\|\vSi_v^{-1}\left[\delta\hat{\vth}_v^2 \; \Delta\hat{\vp}_v^2\right]^\top\right\|_\infty > k^2$.
\section{Simulation Environment}
\begin{figure}[htb]
\centering
\includegraphics[width=\linewidth]{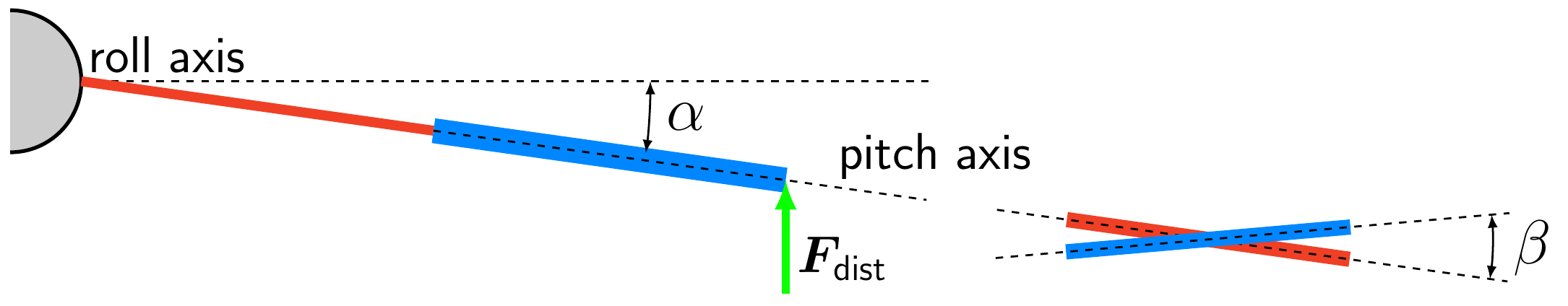}
\caption{View from back and right on the flexible
wing modeled by two rigid bodies.
There are two degrees of freedom: the roll angle $\alpha$ and the pitch angle $\beta$.}
\label{fig:wing}
\end{figure}

\emph{Gazebo}-based \textit{RotorS} \cite{furrer2016rotors} is used to simulate the forces acting on the flexible wings, and to collect precise ground truth poses and IMU measurements.
The camera poses and a publicly available mesh are imported into \textit{Blender} and the photo-realistic images are rendered.
\begin{figure}[htb]
	\includegraphics[width=\textwidth]{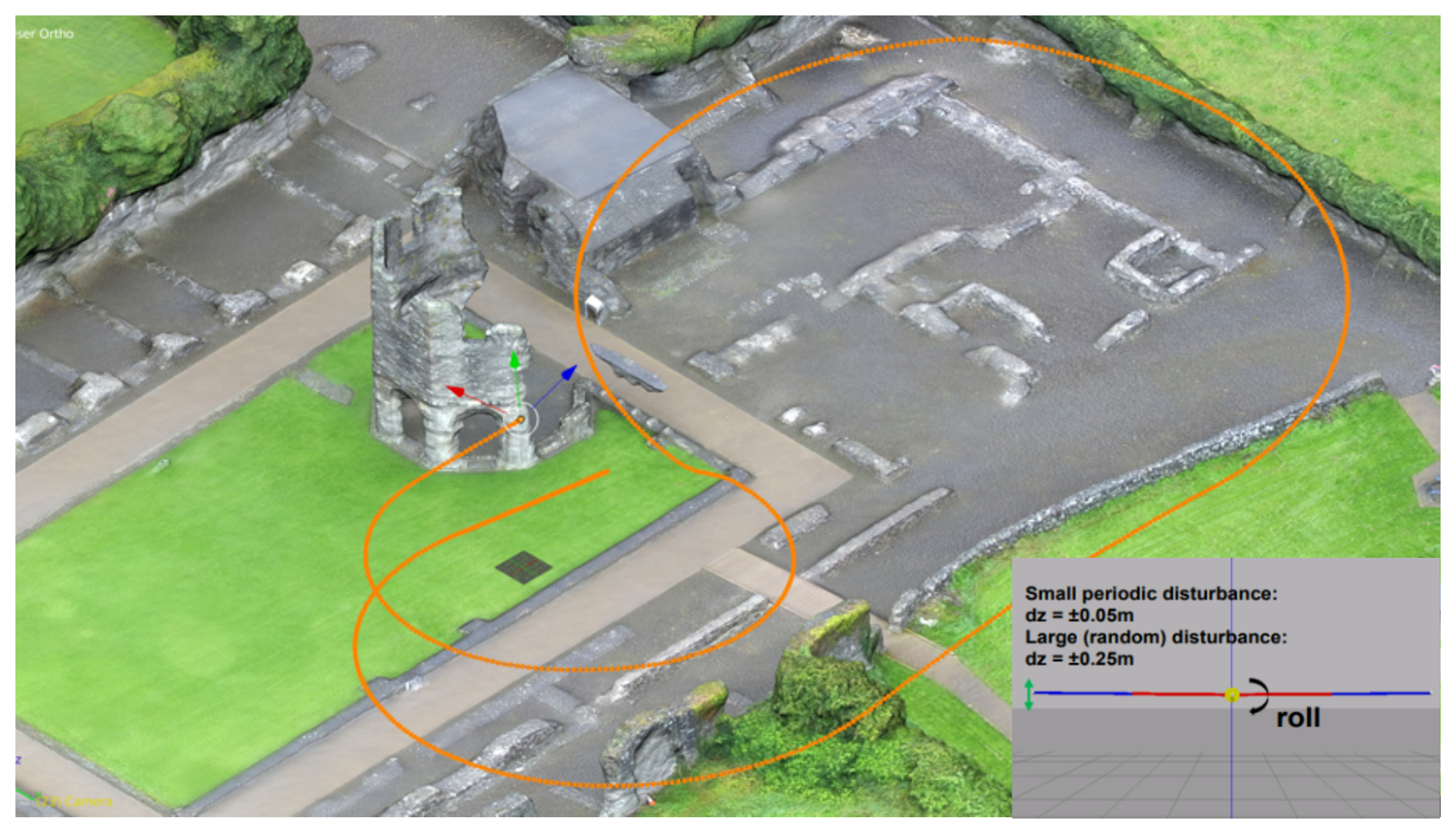}
\centering
\caption{Synthetic dataset: Overview mesh and aircraft trajectory visualized in \emph{Blender}. The image in the bottom right corner shows the aircraft with two flexible wings modeled in \emph{Rotors}. The mesh was downloaded from \url{https://skfb.ly/Sq7J}.}
\label{fig:world}
\end{figure}
\paragraph{Aircraft with flexible wings} The aircraft is modeled by two wings, each connected to the airframe by a joint.
The joints permit only roll motion, \ie rotational movement around the aircraft's body's $\vx$-axis.
Furthermore, as shown in Fig. \ref{fig:wing}, each wing in itself is modeled by two rigid bodies connected by another joint permitting only movement around the wing's center line, \ie pitching movements.
Both joint angles are controlled by a PD-controller which emulates a spring-damper system.
The total mass of the aircraft is \SI{2.8}{kg}, each wing has a mass of \SI{0.4}{kg}.
A disturbance force $\vF_\text{dist} = \vF_p + \vF_r$ is applied on the wing tips.
The disturbance force is composed of a periodic force $\vF_p = a_p \sin\left(f_p t\right)$ with magnitude $a_p$ and frequency $f_p$ and a random force $\vF_r$ of random magnitude $a_r \sim\mathcal{N}(a_r, \sigma_{a_r})$ applied at fixed frequency $f_r$ for a duration of $t_r$.
The random forces simulate the effect of wind gusts and excite the wing into brief non-uniform oscillations.
The sinusoidal force is applied in same phase with a magnitude of $\SI{0.25}{N}$ and frequency of $f_p = \SI{1.5}{Hz}$ on both wing tips.
The random magnitude is sampled from $\mathcal{N}(1.0, 0.1)$ every $\SI{8}{s}$ (i.e. $f_r = \SI{0.125}{Hz}$), however, independently for each side and applied over a duration of $\SI{0.4}{s}$.
The aircraft follows the predefined trajectory as depicted in Fig. \ref{fig:world}.
A 6-DoF PID-controller is used which applies forces along and torques around all three body axes based on the desired velocity and position.
\paragraph{Visual-Inertial Camera Rig}
\begin{figure}[tb]
\raggedleft
\setlength\figureheight{6em}
\setlength\figurewidth{\linewidth - \widthof{$-10$} - 20pt}
\input{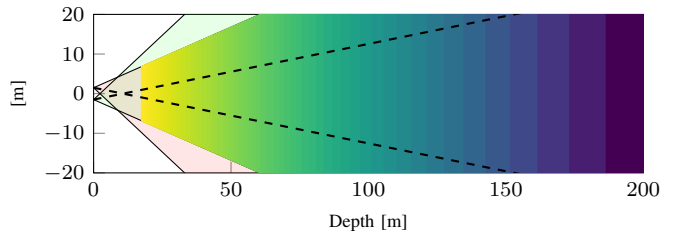}
\caption{The FOV in the $x$-$z$-plane of the left (red) and right (green) camera with a baseline of \SI{3}{m}. 
In the overlapping FOV the area corresponding to each disparity value (12 to 140) is colorized uniformly.}
\label{fig:stereo-rig}
\end{figure}
The geometry of the visual-inertial stereo rig with a nominal baseline of \SI{3}{m} is depicted in Fig. \ref{fig:stereo-rig}.
The cameras\footnote{Perfect pinhole cameras without distortion and a resolution of $720 \times 480$ pixels ($\unit[0.36]{MP}$).} are rotated towards each other by \SI{8}{\degree} to increase the overlapping field of view (FOV). 
The IMU coordinate frames coincide with the camera frames, their axes are kept parallel.
The IMU \textit{ADIS 16448} \cite{adisIMU} is simulated with white noise variances of \SI{1.225e-7}{rad/s} for angular velocities and \SI{1.6e-5}{m^2/s} for linear accelerations.
\paragraph{Initialization of EKF} 
The baseline prior is obtained by empirically estimating the mean and variances of the baseline transform as described in Section \ref{baseline-prior}.
To simulate an imperfect baseline-prior calibration the variances are increased by $\unit[10]{\%}$.
The mean and standard errors of the real relative transform are shown in Table \ref{table:gt}. The mean rotation is parametrized as roll-pitch-yaw angles $\vmu_{\bm{\theta}}$.
\begin{table}[htb]
\footnotesize
\begin{tabular}{|l|l|l|l|l|}\hline
\rowcolor{black!10} &  $\vx$ & $\vy$ & $\vz$ & unit \\ \hline
$\vmu_{\bm{\theta}}$          & -0.51   & -1.8e-6 & -3.3e-4 & deg \\ \hline
$\vsi_{\delta\vth}$ &   1.9    &  7.1e-3 &  1.3e-2 & deg \\ \hline
$\vmu_{\bm{p}}$           &-0.0087 & -3000.0 &  13.4  & mm\\ \hline
$\vsi_{\Delta\vp}$  &   0.27   &  3.0    &  50.5  & mm\\ \hline
\end{tabular}
\caption{True nominal baseline transform used in the synthetic dataset.}
\label{table:gt}
\end{table}
\section{Simulation Results}
\subsection{Evaluation of Estimated Camera Poses}
In this section, we compare the camera pose estimates obtained from a) using the fixed calibration only, b) using joint prior-visual estimates only (with five-point Nister algorithm), c) using the prior calibration and IMU measurements only and finally e) using the full pipeline (prior, vision, and IMU).
Note that the EKF fused with IMU measurements only (without baseline prior) diverges quickly and is not shown in the following.
Fig. \ref{fig:box} visualizes the individual normalized root mean squared errors (RMSE) with respect to the ground truth poses.
There are two main observations: Firstly, incorporating the IMU decreases the errors significantly.
Secondly, since the RMSE of the relative PNP solution is relatively high, fusing them into the EKF reduces the overall RMSE only slightly.
\begin{figure}[htb]
\raggedleft
\setlength\figureheight{1.6cm}
\setlength\figurewidth{\linewidth - \widthof{visual + Prior} }
\begin{tikzpicture}

\definecolor{color1}{rgb}{0.9,0.9,0.9}
\definecolor{color0}{rgb}{1.0,1.0,1.0}
\definecolor{color2}{rgb}{1.0,1.0,1.0}
\definecolor{color3}{rgb}{1.0,1.0,1.0}
\definecolor{median}{HTML}{D62728}
\definecolor{mean}{HTML}{1F77B4}
\tikzstyle{dashed}= [dash pattern=on 2pt off 1pt]

\begin{axis}[
xmin=0.0, xmax=1.25,
ymin=-0.6, ymax=3.6,
width=\figurewidth,
height=\figureheight,
ytick={0,1,2,3},
yticklabels={Full, IMU + Prior, Visual + Prior, Fixed},
xtick align=inside,
ytick align=inside,
tick pos=left,
xmajorgrids,
x grid style={white!80.0!black},
y grid style={white!80.0!black},
axis line style={lightgray!20.0!black},
xlabel near ticks,
ylabel near ticks,
label style={font=\scriptsize},
tick label style={font=\footnotesize},
ylabel style={draw=none},
enlargelimits=false,
scale only axis
]
\path [line width=0.64pt, draw=black, fill=color0] (axis cs:0.0771683571348395,-0.4)
--(axis cs:0.0771683571348395,0.4)
--(axis cs:0.115420441791292,0.4)
--(axis cs:0.124075814767088,0.2)
--(axis cs:0.132731187742884,0.4)
--(axis cs:0.233098877514054,0.4)
--(axis cs:0.233098877514054,-0.4)
--(axis cs:0.132731187742884,-0.4)
--(axis cs:0.124075814767088,-0.2)
--(axis cs:0.115420441791292,-0.4)
--(axis cs:0.0771683571348395,-0.4)
--cycle;

\path [line width=0.64pt, draw=black, fill=color1] (axis cs:0.0770929555376533,0.6)
--(axis cs:0.0770929555376533,1.4)
--(axis cs:0.119223934697325,1.4)
--(axis cs:0.127885159589615,1.2)
--(axis cs:0.136546384481905,1.4)
--(axis cs:0.233128900880838,1.4)
--(axis cs:0.233128900880838,0.6)
--(axis cs:0.136546384481905,0.6)
--(axis cs:0.127885159589615,0.8)
--(axis cs:0.119223934697325,0.6)
--(axis cs:0.0770929555376533,0.6)
--cycle;

\path [line width=0.64pt, draw=black, fill=color2] (axis cs:0.298162444953742,1.6)
--(axis cs:0.298162444953742,2.4)
--(axis cs:0.542261301596103,2.4)
--(axis cs:0.577755166135215,2.2)
--(axis cs:0.613249030674327,2.4)
--(axis cs:0.937600720268974,2.4)
--(axis cs:0.937600720268974,1.6)
--(axis cs:0.613249030674327,1.6)
--(axis cs:0.577755166135215,1.8)
--(axis cs:0.542261301596103,1.6)
--(axis cs:0.298162444953742,1.6)
--cycle;

\path [line width=0.64pt, draw=black, fill=color3] (axis cs:0.340783224223767,2.6)
--(axis cs:0.340783224223767,3.4)
--(axis cs:0.626596349712634,3.4)
--(axis cs:0.667431767162882,3.2)
--(axis cs:0.70826718461313,3.4)
--(axis cs:1.07645202910666,3.4)
--(axis cs:1.07645202910666,2.6)
--(axis cs:0.70826718461313,2.6)
--(axis cs:0.667431767162882,2.8)
--(axis cs:0.626596349712634,2.6)
--(axis cs:0.340783224223767,2.6)
--cycle;

\addplot [line width=0.64pt, black, forget plot]
table {%
0.0771683571348395 0
0.00646623483979438 0
};
\addplot [line width=0.64pt, black, forget plot]
table {%
0.233098877514054 0
0.46554139019212 0
};
\addplot [line width=0.64pt, black, forget plot]
table {%
0.00646623483979438 -0.2
0.00646623483979438 0.2
};
\addplot [line width=0.64pt, black, forget plot]
table {%
0.46554139019212 -0.2
0.46554139019212 0.2
};
\addplot [line width=0.64pt, black, forget plot]
table {%
0.0770929555376533 1
0.00753835804841952 1
};
\addplot [line width=0.64pt, black, forget plot]
table {%
0.233128900880838 1
0.464688391434457 1
};
\addplot [line width=0.64pt, black, forget plot]
table {%
0.00753835804841952 0.8
0.00753835804841952 1.2
};
\addplot [line width=0.64pt, black, forget plot]
table {%
0.464688391434457 0.8
0.464688391434457 1.2
};
\addplot [line width=0.64pt, black, forget plot]
table {%
0.298162444953742 2
0.0165277163771368 2
};
\addplot [line width=0.64pt, black, forget plot]
table {%
0.937600720268974 2
1.87072439768876 2
};
\addplot [line width=0.64pt, black, forget plot]
table {%
0.0165277163771368 1.8
0.0165277163771368 2.2
};
\addplot [line width=0.64pt, black, forget plot]
table {%
1.87072439768876 1.8
1.87072439768876 2.2
};
\addplot [line width=0.64pt, black, forget plot]
table {%
0.340783224223767 3
0.0165285583995513 3
};
\addplot [line width=0.64pt, black, forget plot]
table {%
1.07645202910666 3
2.17818536966014 3
};
\addplot [line width=0.64pt, black, forget plot]
table {%
0.0165285583995513 2.8
0.0165285583995513 3.2
};
\addplot [line width=0.64pt, black, forget plot]
table {%
2.17818536966014 2.8
2.17818536966014 3.2
};
\addplot [line width=0.64pt, median, forget plot]
table {%
0.124075814767088 -0.2
0.124075814767088 0.2
};
\addplot [thick, mean, dashed, forget plot]
table {%
0.192282570504009 -0.4
0.192282570504009 0.4
};
\addplot [line width=0.64pt, median, forget plot]
table {%
0.127885159589615 0.8
0.127885159589615 1.2
};
\addplot [thick, mean, dashed, forget plot]
table {%
0.19384127875215 0.6
0.19384127875215 1.4
};
\addplot [line width=0.64pt, median, forget plot]
table {%
0.577755166135215 1.8
0.577755166135215 2.2
};
\addplot [thick, mean, dashed, forget plot]
table {%
0.816489300326353 1.6
0.816489300326353 2.4
};
\addplot [line width=0.64pt, median, forget plot]
table {%
0.667431767162882 2.8
0.667431767162882 3.2
};
\addplot [thick, mean, dashed, forget plot]
table {%
0.906045352862234 2.6
0.906045352862234 3.4
};
\end{axis}
\end{tikzpicture}
\caption{Normalized RMSE of a fixed baseline assumption, fused estimate of visual pose estimation with the baseline prior, using IMU and baseline prior, and the full pipeline (visual, IMU, and baseline prior).
The dashed line shows the mean.}
\label{fig:box}
\end{figure}
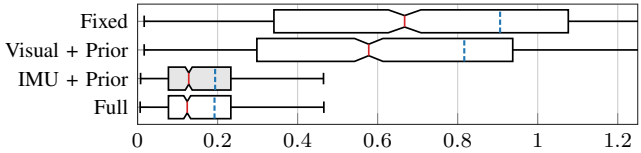
\begin{table}[htb]
\footnotesize
\centering 
\begin{tabular}{|l|l|l|l|l|l|l|} \hline
\rowcolor{black!10}  &$\delta\tilde\vth_x$ & $\delta\tilde\vth_y$ & $\delta\tilde\vth_z$
& $\Delta\tilde\vp_x$ & $\Delta\tilde\vp_y$ & $\Delta\tilde\vp_z$  \\ \hline
Fixed    & 1.96  & 0.0071 & 0.0102 & 0.269 & 3.06   & 51.2 \\ \hline 
IMU + Prior & 0.083 & 0.0070 & 0.0095 & 0.375 & 2.83   & 14.7 \\ \hline
\end{tabular}
\caption{Rotational RMSE in $\unit[]{deg}$ and translational RMSE in $\unit[]{mm}$.}
\label{table:est}
\end{table}
\normalsize
As can be seen in Table \ref{table:est}, the rotations around the $\vy$- (pitch) and $\vz$-axis (yaw) as well as the motions along $\vx$- and $\vy$-axis have similarly small deviations as our error estimate and hence do not significantly affect correspondence matching or depth estimation.
For the rotation around $\vx$-axis we see an improvement of over one order of magnitude (from around \SI{2}{\degree} to \SI{0.08}{\degree}) and in movement along $\vz$-axis a three-fold improvement.
Rotational disturbances affect the matching process much stronger than translational disturbances, especially since we deal with objects far away.
Given our large baseline and long distances, the translational deviations on $\vy$- and $\vz$-axis do not have a significant impact.
For instance, the disturbance of around \SI{15}{mm} on $\vz$-axis, which corresponds to the optical $\vy$ axis, does not result in any pixel disturbance for points that are \SI{13}{m} or further away.
%
%
\begin{figure}[tb]
\raggedleft
\setlength\figureheight{2cm}
\setlength\figurewidth{\linewidth - 2.8em}
\input{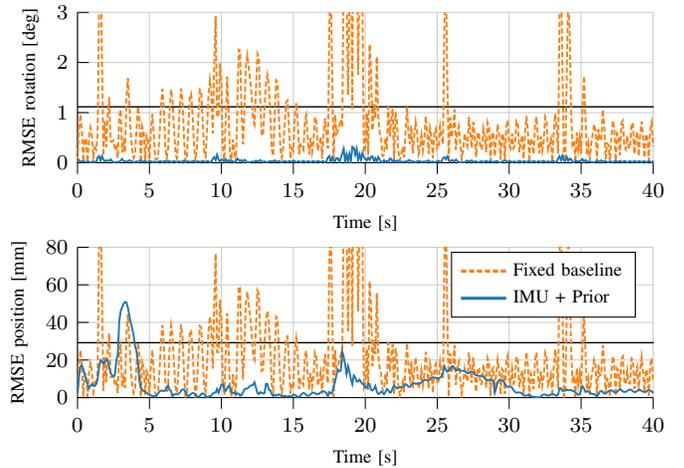}
\caption{RMS errors plotted over the first \SI{40}{s} in orientation and position of estimated deviations based on the fixed baseline assumption or using the proposed pipeline.
The random disturbance force results in high peaks every \SI{8}{s} while the periodic disturbance results in high-frequent errors.
The horizontal black line represents the standard error of the disturbance.}
\label{fig:est-time}
\end{figure}
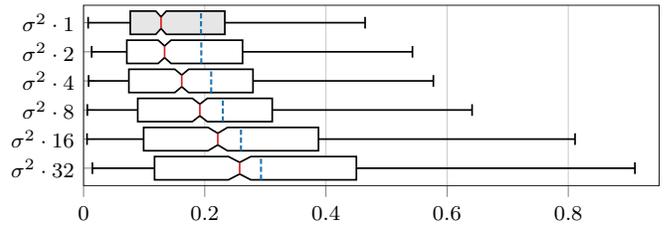
\begin{figure}[htb]
\raggedleft
\setlength\figurewidth{\linewidth - 2.8em}
\setlength\figureheight{2.4cm}
\begin{tikzpicture}
\definecolor{color0}{rgb}{1.0,1.0,1.0}
\definecolor{color1}{rgb}{1.0,1.0,1.0}
\definecolor{color2}{rgb}{1.0,1.0,1.0}
\definecolor{color3}{rgb}{1.0,1.0,1.0}
\definecolor{color4}{rgb}{1.0,1.0,1.0}
\definecolor{color5}{rgb}{0.9,0.9,0.9}
\definecolor{median}{HTML}{D62728}
\definecolor{mean}{HTML}{1F77B4}
\tikzstyle{dashed}= [dash pattern=on 2pt off 1pt]
\begin{axis}[
xmin=0.0, xmax=0.95,
ymin=-0.6, ymax=5.6,
width=\figurewidth,
height=\figureheight,
ytick={0,1,2,3,4,5},
yticklabels={$\sigma^2 \cdot 32$, $\sigma^2 \cdot 16$, $\sigma^2 \cdot 8$, $\sigma^2 \cdot 4$, $\sigma^2 \cdot 2$, $\sigma^2 \cdot 1$},
xtick align=outside,
ytick align=inside,
tick pos=left,
xmajorgrids,
x grid style={white!80.0!black},
y grid style={white!80.0!black},
axis line style={lightgray!20.0!black},
xlabel near ticks,
ylabel near ticks,
label style={font=\scriptsize},
tick label style={font=\footnotesize},
enlargelimits=false,
scale only axis
]
\path [line width=0.64pt, draw=black, fill=color0] (axis cs:0.117100122408627,-0.4)
--(axis cs:0.117100122408627,0.4)
--(axis cs:0.239143477489225,0.4)
--(axis cs:0.257636759030286,0.2)
--(axis cs:0.276130040571347,0.4)
--(axis cs:0.450265085060249,0.4)
--(axis cs:0.450265085060249,-0.4)
--(axis cs:0.276130040571347,-0.4)
--(axis cs:0.257636759030286,-0.2)
--(axis cs:0.239143477489225,-0.4)
--(axis cs:0.117100122408627,-0.4)
--cycle;
\path [line width=0.64pt, draw=black, fill=color1] (axis cs:0.0991105715346002,0.6)
--(axis cs:0.0991105715346002,1.4)
--(axis cs:0.205506569500354,1.4)
--(axis cs:0.221526648606271,1.2)
--(axis cs:0.237546727712189,1.4)
--(axis cs:0.387719656144522,1.4)
--(axis cs:0.387719656144522,0.6)
--(axis cs:0.237546727712189,0.6)
--(axis cs:0.221526648606271,0.8)
--(axis cs:0.205506569500354,0.6)
--(axis cs:0.0991105715346002,0.6)
--cycle;
\path [line width=0.64pt, draw=black, fill=color2] (axis cs:0.0893972732224006,1.6)
--(axis cs:0.0893972732224006,2.4)
--(axis cs:0.179606464816908,2.4)
--(axis cs:0.191929218893726,2.2)
--(axis cs:0.204251972970544,2.4)
--(axis cs:0.31139734890679,2.4)
--(axis cs:0.31139734890679,1.6)
--(axis cs:0.204251972970544,1.6)
--(axis cs:0.191929218893726,1.8)
--(axis cs:0.179606464816908,1.6)
--(axis cs:0.0893972732224006,1.6)
--cycle;
\path [line width=0.64pt, draw=black, fill=color3] (axis cs:0.074751983105334,2.6)
--(axis cs:0.074751983105334,3.4)
--(axis cs:0.15066482324259,3.4)
--(axis cs:0.162038314164839,3.2)
--(axis cs:0.173411805087087,3.4)
--(axis cs:0.279650646975036,3.4)
--(axis cs:0.279650646975036,2.6)
--(axis cs:0.173411805087087,2.6)
--(axis cs:0.162038314164839,2.8)
--(axis cs:0.15066482324259,2.6)
--(axis cs:0.074751983105334,2.6)
--cycle;
\path [line width=0.64pt, draw=black, fill=color4] (axis cs:0.071364427182759,3.6)
--(axis cs:0.071364427182759,4.4)
--(axis cs:0.123113098177948,4.4)
--(axis cs:0.133717899607103,4.2)
--(axis cs:0.144322701036258,4.4)
--(axis cs:0.262414796703577,4.4)
--(axis cs:0.262414796703577,3.6)
--(axis cs:0.144322701036258,3.6)
--(axis cs:0.133717899607103,3.8)
--(axis cs:0.123113098177948,3.6)
--(axis cs:0.071364427182759,3.6)
--cycle;
\path [line width=0.64pt, draw=black, fill=color5] (axis cs:0.0770929555376533,4.6)
--(axis cs:0.0770929555376533,5.4)
--(axis cs:0.119223934697325,5.4)
--(axis cs:0.127885159589615,5.2)
--(axis cs:0.136546384481905,5.4)
--(axis cs:0.233128900880838,5.4)
--(axis cs:0.233128900880838,4.6)
--(axis cs:0.136546384481905,4.6)
--(axis cs:0.127885159589615,4.8)
--(axis cs:0.119223934697325,4.6)
--(axis cs:0.0770929555376533,4.6)
--cycle;
\addplot [line width=0.64pt, black, forget plot]
table {%
0.117100122408627 0
0.014590209067146 0
};
\addplot [line width=0.64pt, black, forget plot]
table {%
0.450265085060249 0
0.910025987627582 0
};
\addplot [line width=0.64pt, black, forget plot]
table {%
0.014590209067146 -0.2
0.014590209067146 0.2
};
\addplot [line width=0.64pt, black, forget plot]
table {%
0.910025987627582 -0.2
0.910025987627582 0.2
};
\addplot [line width=0.64pt, black, forget plot]
table {%
0.0991105715346002 1
0.00569208311943842 1
};
\addplot [line width=0.64pt, black, forget plot]
table {%
0.387719656144522 1
0.811209093492004 1
};
\addplot [line width=0.64pt, black, forget plot]
table {%
0.00569208311943842 0.8
0.00569208311943842 1.2
};
\addplot [line width=0.64pt, black, forget plot]
table {%
0.811209093492004 0.8
0.811209093492004 1.2
};
\addplot [line width=0.64pt, black, forget plot]
table {%
0.0893972732224006 2
0.00626601478777859 2
};
\addplot [line width=0.64pt, black, forget plot]
table {%
0.31139734890679 2
0.64127932732151 2
};
\addplot [line width=0.64pt, black, forget plot]
table {%
0.00626601478777859 1.8
0.00626601478777859 2.2
};
\addplot [line width=0.64pt, black, forget plot]
table {%
0.64127932732151 1.8
0.64127932732151 2.2
};
\addplot [line width=0.64pt, black, forget plot]
table {%
0.074751983105334 3
0.00827739480631172 3
};
\addplot [line width=0.64pt, black, forget plot]
table {%
0.279650646975036 3
0.57736953518468 3
};
\addplot [line width=0.64pt, black, forget plot]
table {%
0.00827739480631172 2.8
0.00827739480631172 3.2
};
\addplot [line width=0.64pt, black, forget plot]
table {%
0.57736953518468 2.8
0.57736953518468 3.2
};
\addplot [line width=0.64pt, black, forget plot]
table {%
0.071364427182759 4
0.0133364305561864 4
};
\addplot [line width=0.64pt, black, forget plot]
table {%
0.262414796703577 4
0.542860192970257 4
};
\addplot [line width=0.64pt, black, forget plot]
table {%
0.0133364305561864 3.8
0.0133364305561864 4.2
};
\addplot [line width=0.64pt, black, forget plot]
table {%
0.542860192970257 3.8
0.542860192970257 4.2
};
\addplot [line width=0.64pt, black, forget plot]
table {%
0.0770929555376533 5
0.00753835804841952 5
};
\addplot [line width=0.64pt, black, forget plot]
table {%
0.233128900880838 5
0.464688391434457 5
};
\addplot [line width=0.64pt, black, forget plot]
table {%
0.00753835804841952 4.8
0.00753835804841952 5.2
};
\addplot [line width=0.64pt, black, forget plot]
table {%
0.464688391434457 4.8
0.464688391434457 5.2
};
\addplot [line width=0.64pt, median, forget plot]
table {%
0.257636759030286 -0.2
0.257636759030286 0.2
};
\addplot [thick, mean, dashed, forget plot]
table {%
0.292868802459516 -0.4
0.292868802459516 0.4
};
\addplot [line width=0.64pt, median, forget plot]
table {%
0.221526648606271 0.8
0.221526648606271 1.2
};
\addplot [thick, mean, dashed, forget plot]
table {%
0.259810281437047 0.6
0.259810281437047 1.4
};
\addplot [line width=0.64pt, median, forget plot]
table {%
0.191929218893726 1.8
0.191929218893726 2.2
};
\addplot [thick, mean, dashed, forget plot]
table {%
0.229889778238047 1.6
0.229889778238047 2.4
};
\addplot [line width=0.64pt, median, forget plot]
table {%
0.162038314164839 2.8
0.162038314164839 3.2
};
\addplot [thick, mean, dashed, forget plot]
table {%
0.210644901710233 2.6
0.210644901710233 3.4
};
\addplot [line width=0.64pt, median, forget plot]
table {%
0.133717899607103 3.8
0.133717899607103 4.2
};
\addplot [thick, mean, dashed, forget plot]
table {%
0.19430266266896 3.6
0.19430266266896 4.4
};
\addplot [line width=0.64pt, median, forget plot]
table {%
0.127885159589615 4.8
0.127885159589615 5.2
};
\addplot [thick, mean, dashed, forget plot]
table {%
0.19384127875215 4.6
0.19384127875215 5.4
};
\end{axis}
\end{tikzpicture}
\caption{Normalized RMSE versus noise-level expressed as multiples of the noise variances of the \textit{ADIS 16448} IMU. Note the logarithmic y-axis.}
\label{fig:noise}
\end{figure}
Comparing to the fixed baseline, which suffers both from the periodic, regular disturbances, and the low-frequent random disturbances, our proposed approach performs significantly better as further shown in Fig. \ref{fig:est-time}.
The first peak at \SI{4}{s} in positional error is due to the initialization of the EKF.
Relative velocity, angular velocities, and linear accelerations are all initialized to zero.
Position and orientation are initialized using the baseline prior.
However since our trajectory starts mid-air, our initialization values for the EKF are incorrect, especially for the linear accelerations.
This leads to relatively bad early estimates in position of up to about \SI{5}{s}, until the acceleration values and thus velocity estimates converge.
This behavior is less pronounced in orientation since the angular velocities are directly measured.
The small peak at around \SI{17}{s} can be explained by an exceptionally large random disturbance at this time step.
Since our IMUs operate at only \SI{100}{Hz} not all acceleration spikes can be captured correctly.
This leads to wrong acceleration values similar to the wrong initialization case. 

In a next step the sensitivity of our approach with respect to IMU noise is analyzed.
Fig. \ref{fig:noise} plots the obtained normalized RMS errors for the proposed pipeline.
We express the noise levels as multiples of the noise of the \textit{ADIS 16448} IMU.
Clearly, increasing noise affects the estimates, but for up to about an eight-fold increase in noise the estimates still stay relatively close.
From this experiment, we conclude that even in real-world scenarios our approach is expected to yield good results.
\subsection{Evaluation of Depth-Maps}
\begin{figure}[tb]
\setlength\figurewidth{\linewidth - 2.8em}
\setlength\figureheight{0.1cm}
\input{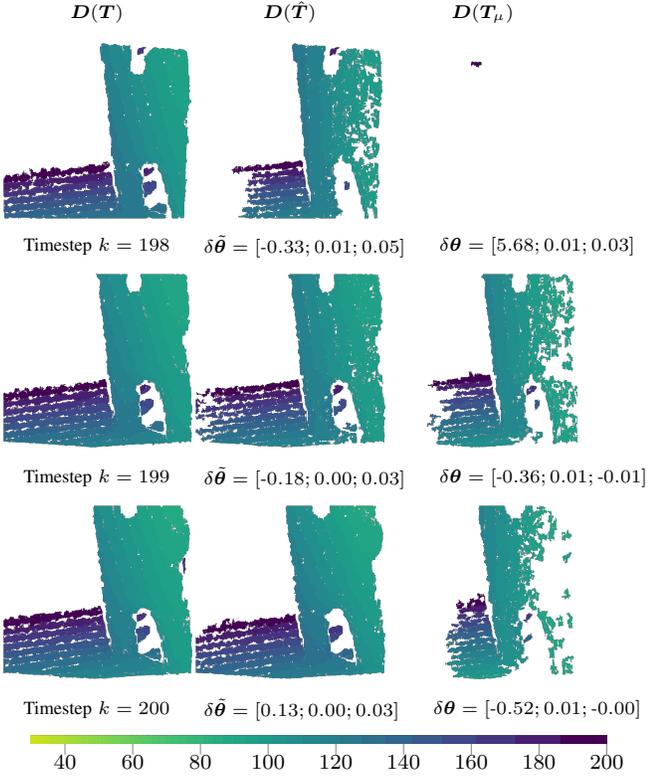}
\caption{%
Timesteps 198--200: Depth maps obtained with $\vT$, $\hat\vT$ and $\vT_\mu$.
The given colormap is truncated from the full range (cf.\ Fig. \ref{fig:stereo-rig}).
Note that the colormap is intentionally not continuous but all depths that are mapped to one disparity value are colored uniformly.
}
\label{fig:depth-maps}
\end{figure}

Finally, we wish to support our claim that the resulting relative pose estimates are of such a high quality that accurate depth maps can be obtained.
For comparison, three depth maps are generated at each timestep: one based on the ground-truth transform $\vT$, one based on our estimate $\hat \vT$, and one based on a fixed baseline transformation, denoted by $\vT_\mu$.
The disparities of dense correspondences in the rectified stereo images are computed using Stereo Block Matching (BM) \cite{bradski2000opencv}.
%
%
%
%
Given the disparity $d$ of a pixel in the rectified images, the depth of a pixel is given by $z = f\cdot b / d$ where $f$ is the focal length and $b$ the baseline of the rectified stereo pair.
The obtained depth map image from transform $\vT$ is denoted by $\vD(\vT)$.
First, the performance of our approach is qualitatively demonstrated based on three characteristic frames in Fig. \ref{fig:depth-maps}.
For each frame color-coded depth maps are shown and denoted by $\vD(\vT)$, $\vD(\hat\vT)$ and $\vD(\vT_\mu)$, \ie depth maps obtained from the real, the estimated respectively the mean baseline transform.
%
%
Let us first focus on what happens with the depth map based on a fixed baseline $\vD(\vT_\mu)$: 
In frame $198$, it is evident that the stereo correspondence search failed almost completely.
This demonstrates exemplary how an error in roll angle $\vth_x$ affects the matching process in such a way that no correspondences are found.
Due to an erroneous rectification, the epipolar lines, along which the correspondences are searched, are flawed.
In timestep $199$, however, one can see that $\vD(\vT_\mu)$ resembles the ground truth depth map.
This is due to the fact that the relative transform periodically ``traverses through the origin'', \ie corresponds to the mean baseline.
In these cases, $\vT_\mu$ is close to $\vT$ and thus errors are small.
In the last frame, at timestep $200$, this effect already is seen to be diminished.
The depth maps obtained using the poses estimated by our proposed framework show only small reconstruction errors compared to the ground-truth (at timestep $200$).
%
In frame $198$, the disturbance is not estimated perfectly, evident by a remaining error in roll of almost \SI{0.4}{\degree}.
This affects mostly pixels lying towards the left and right of the image due to the correspondence search along epipolar lines.
This can also be observed in depth map $\vD(\vT_\mu)$ in timestep $199$.
In contrast, for pixels close to the center line a slight misalignment does not result in a matching failure.
%
%

There are two different errors present in the generated depth maps.
We consider pixels that are valid in the ground-truth depth map.
For each valid pixel in $\Delta(\vT)$, two error cases can occur:
First, this pixel might be invalid in the estimated depth map.
Second, the pixel might be valid, but off by a certain value.
In order to quantify these errors, the error functions $\Delta_\#(\vD, \hat\vD)$, $\Delta_z(\vD, \hat\vD)$ are defined.
$\Delta_\#(\vD, \hat\vD)$ is defined to be the fraction of additionally invalid pixels in the estimate $\hat\vD$ compared to the total number of valid pixels (in $\vD$), i.e.\ capturing the \emph{completeness} of the depth map estimate. 
$\Delta_z(\vD,\hat\vD)$ is defined to be the RMS error in depth for all valid pixels thus a measure of \emph{accuracy} for the estimates.
Note that pixels in $\hat\vD$ are ignored even if they are valid in $\hat\vD$ but not valid in $\vD$.
Defining the indicator function $v(\vD_{i,j}) = [\vD_{i,j} \text{is valid}]$ and $\delta(\vD_{i,j}, \hat\vD_{i,j}) = \vD_{i,j} - \hat\vD_{i,j}$ if both pixels are valid and else $0$, the formula to calculate these two errors is given by:
\begin{align}
	\Delta_\#(\vD, \hat\vD) &= \left(\frac{\sum_{i,j} [v(\vD_{i,j}) \wedge \neg v(\hat\vD_{i, j})]}{\sum_{i,j} v(\vD_{i,j})}\right)^{1/2} \\
	\Delta_z(\vD, \hat\vD) &= \left(\sum_{i,j} \delta^2(\vD_{i,j} - \hat\vD_{i,j})\right)^{1/2}.
\end{align}
\begin{figure}[htb]
\setlength\figureheight{0.8cm}
\setlength{\figurewidth}{\linewidth - \widthof{$D(\vT_\mu)$} - 5pt}
\begin{tikzpicture}

\definecolor{color2}{HTML}{FFFFFF}
\definecolor{color1}{rgb}{0.9,0.9,0.9}
\definecolor{median}{HTML}{D62728}
\definecolor{mean}{HTML}{1F77B4}
\tikzstyle{dashed}= [dash pattern=on 2pt off 1pt]

\begin{groupplot}[group style={group size=1 by 2, vertical sep=4em}]
\nextgroupplot[
title={\footnotesize Accuracy $\Delta_z(\vD,\hat\vD)$ [m]},
xmin=0, xmax=28,
ymin=-0.6, ymax=1.6,
width=\figurewidth,
height=\figureheight,
ytick={0,1},
yticklabels={$\vD(\hat\vT)$,$\vD(\vT_\mu)$},
tick align=inside,
tick pos=left,
x grid style={white!80.0!black},
y grid style={white!80.0!black},
axis line style={lightgray!20.0!black},
label style={font=\scriptsize},
tick label style={font=\footnotesize},
enlargelimits=false,
scale only axis
]

\path [line width=0.72pt, draw=black, fill=color1] (axis cs:0.833136928490526,-0.4)
--(axis cs:0.833136928490526,0.4)
--(axis cs:1.3753111779237,0.4)
--(axis cs:1.48457066695942,0.2)
--(axis cs:1.59383015599515,0.4)
--(axis cs:2.801496816545,0.4)
--(axis cs:2.801496816545,-0.4)
--(axis cs:1.59383015599515,-0.4)
--(axis cs:1.48457066695942,-0.2)
--(axis cs:1.3753111779237,-0.4)
--(axis cs:0.833136928490526,-0.4)
--cycle;

\path [line width=0.72pt, draw=black, fill=color2] (axis cs:1.20815746914677,0.6)
--(axis cs:1.20815746914677,1.4)
--(axis cs:7.15694056999393,1.4)
--(axis cs:8.50630262454029,1.2)
--(axis cs:9.85566467908664,1.4)
--(axis cs:25.5175347696701,1.4)
--(axis cs:25.5175347696701,0.6)
--(axis cs:9.85566467908664,0.6)
--(axis cs:8.50630262454029,0.8)
--(axis cs:7.15694056999393,0.6)
--(axis cs:1.20815746914677,0.6)
--cycle;

\addplot [line width=0.72pt, black, forget plot]
table {%
0.833136928490526 0
0 0
};
\addplot [line width=0.72pt, black, forget plot]
table {%
2.801496816545 0
5.65299924798646 0
};
\addplot [line width=0.72pt, black, forget plot]
table {%
0 -0.2
0 0.2
};
\addplot [line width=0.72pt, black, forget plot]
table {%
5.65299924798646 -0.2
5.65299924798646 0.2
};
\addplot [line width=0.72pt, black, forget plot]
table {%
1.20815746914677 1
0 1
};
\addplot [line width=0.72pt, black, forget plot]
table {%
25.5175347696701 1
60.4789087127504 1
};
\addplot [line width=0.72pt, black, forget plot]
table {%
0 0.8
0 1.2
};
\addplot [line width=0.72pt, black, forget plot]
table {%
60.4789087127504 0.8
60.4789087127504 1.2
};
\addplot [line width=0.72pt, median, forget plot]
table {%
1.48457066695942 -0.2
1.48457066695942 0.2
};
\addplot [thick, mean, dashed, forget plot]
table {%
2.17806127591541 -0.4
2.17806127591541 0.4
};
\addplot [line width=0.72pt, median, forget plot]
table {%
8.50630262454029 0.8
8.50630262454029 1.2
};
\addplot [thick, mean, dashed, forget plot]
table {%
16.0699262053834 0.6
16.0699262053834 1.4
};
\nextgroupplot[
xmin=0, xmax=105,
ymin=-0.6, ymax=1.6,
width=\figurewidth,
height=\figureheight,
ytick={0,1},
yticklabels={$\vD(\hat\vT)$,$\vD(\vT_\mu)$},
xticklabel={\pgfmathparse{\tick}\pgfmathprintnumber{\pgfmathresult}\%},
tick align=inside,
tick pos=left,
x grid style={white!80.0!black},
y grid style={white!80.0!black},
axis line style={lightgray!20.0!black},
label style={font=\scriptsize},
tick label style={font=\footnotesize},
enlargelimits=false,
scale only axis,
title={\footnotesize Completeness $\Delta_\#(\vD,\hat\vD)$}
]
\path [line width=0.72pt, draw=black, fill=color1] (axis cs:3.60321472344878,-0.4)
--(axis cs:3.60321472344878,0.4)
--(axis cs:6.70852035859318,0.4)
--(axis cs:7.25710266371244,0.2)
--(axis cs:7.80568496883171,0.4)
--(axis cs:13.4861769652012,0.4)
--(axis cs:13.4861769652012,-0.4)
--(axis cs:7.80568496883171,-0.4)
--(axis cs:7.25710266371244,-0.2)
--(axis cs:6.70852035859318,-0.4)
--(axis cs:3.60321472344878,-0.4)
--cycle;

\path [line width=0.72pt, draw=black, fill=color2] (axis cs:63.4742151460011,0.6)
--(axis cs:63.4742151460011,1.4)
--(axis cs:82.4404441202804,1.4)
--(axis cs:84.2969808132925,1.2)
--(axis cs:86.1535175063046,1.4)
--(axis cs:96.9205765511091,1.4)
--(axis cs:96.9205765511091,0.6)
--(axis cs:86.1535175063046,0.6)
--(axis cs:84.2969808132925,0.8)
--(axis cs:82.4404441202804,0.6)
--(axis cs:63.4742151460011,0.6)
--cycle;

\addplot [line width=0.72pt, black, forget plot]
table {%
3.60321472344878 0
0 0
};
\addplot [line width=0.72pt, black, forget plot]
table {%
13.4861769652012 0
28.0731423739562 0
};
\addplot [line width=0.72pt, black, forget plot]
table {%
0 -0.2
0 0.2
};
\addplot [line width=0.72pt, black, forget plot]
table {%
28.0731423739562 -0.2
28.0731423739562 0.2
};
\addplot [line width=0.72pt, black, forget plot]
table {%
63.4742151460011 1
13.787809634455 1
};
\addplot [line width=0.72pt, black, forget plot]
table {%
96.9205765511091 1
100 1
};
\addplot [line width=0.72pt, black, forget plot]
table {%
13.787809634455 0.8
13.787809634455 1.2
};
\addplot [line width=0.72pt, black, forget plot]
table {%
100 0.8
100 1.2
};
\addplot [line width=0.72pt, median, forget plot]
table {%
7.25710266371244 -0.2
7.25710266371244 0.2
};
\addplot [thick, mean, dashed, forget plot]
table {%
11.4607908906643 -0.4
11.4607908906643 0.4
};
\addplot [line width=0.72pt, median, forget plot]
table {%
84.2969808132925 0.8
84.2969808132925 1.2
};
\addplot [thick, mean, dashed, forget plot]
table {%
73.7911037893968 0.6
73.7911037893968 1.4
};
\end{groupplot}

\end{tikzpicture}
\caption{Comparing the errors in accuracy (top) and completeness (bottom).}
\label{fig:depth-box}
\end{figure}
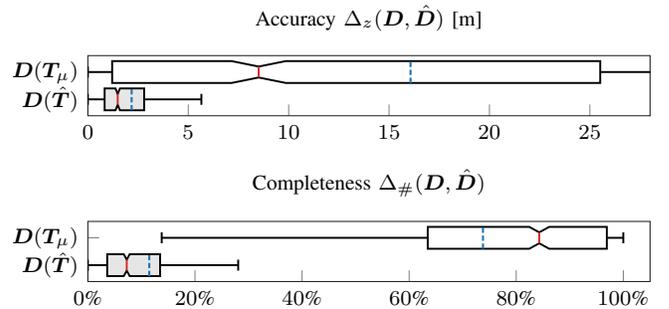
These two error functions are evaluated at each timestep and the result is aggregated in form of box plots as shown in Fig. \ref{fig:depth-box}.
The improvement over the fixed baseline assumption is drastic.
From a mean of over \SI{16}{m} average depth error, the proposed framework reduces the depth error to a mere \SI{2.2}{m}.
For comparison, the average scene depth over all valid ground-truth pixels is \SI{122}{m}.
In completeness, a similar improvement is demonstrated:
On average, almost $\unit[74]{\%}$ of all pixels that are valid in the ground-truth depth map $\vD$ are invalid in $\vD_\mu$.
For depth maps obtained using the proposed framework, however, this number is reduced to an average of $\unit[11.5]{\%}$.
As stated above and observable in Fig. \ref{fig:depth-maps}, these invalid pixels occur mostly towards the edge of the images.
However, as our goal is to detect and avoid obstacles, the important region of interest lies in the middle of the image.
\section{CONCLUSIONS}
In this paper, we present the theory to accurately estimate the time-varying baseline transformation of a flexible wide-baseline stereo pair in order to generate high-quality depth maps.
The light-weight nature of the proposed EKF, as well as the extensive analysis of the relative pose estimates and depth map led to promising results.
In particular, including the measurements of the two rigidly attached IMUs resulted in a significant reduction of the baseline transformation error. 
While the two IMUs fused with an EKF precisely estimate the deformation of the relative baseline transformation in the short-term, the Gaussian prior for the wing model ensures a constrained estimation problem in the long-term.
The incorporation of more sophisticated wing models could further improve the result of the relative baseline estimate.
For instance, a long short-term memory (LSTM) network can learn the periodic wing deflection and predict the future relative poses as a time series.
Extensive experiments need to show if a similar level of robustness, precision, and accuracy is also achievable in the real-world by handling camera and IMU time synchronization and IMU bias estimation.
%
%
\section*{ACKNOWLEDGMENT}
The research leading to these results has received funding from the \emph{Federal office armasuisse Science and Technology} under project n\textdegree 050-45.
Furthermore, we wish to thank Lucas Teixeira (Vision for Robotics Lab, ETH Zurich) for providing scripts that bridge the gap between Blender and Gazebo.
%
%
\bibliographystyle{ieeetr}
\bibliography{lib.bib}
\end{document}